\documentclass[runningheads]{llncs}
\usepackage{blocked}
\usepackage{color}
\usepackage{multirow}
\usepackage{graphicx}
\usepackage{booktabs}
\begin{document}
%
\title{Data-driven analysis of central bank digital currency (CBDC) projects drivers}
%
%
\author{Toshiko Matsui \and
Daniel Perez}
\authorrunning{T. Matsui et al.}
%
\institute{Department of Computing, Imperial College London, United Kingdom\\
\email{\{t.matsui19, daniel.perez\}@imperial.ac.uk}}
\maketitle              
\begin{abstract}
In this paper, we use a variety of machine learning methods to quantify the extent to which economic and technological factors are predictive of the progression of Central Bank Digital Currencies (CBDC) within a country, using as our measure of this progression the CBDC project index (CBDCPI).
We find that a financial development index is the most important feature for our model, followed by the GDP per capita and an index of the voice and accountability of the country’s population. 
Our results are consistent with previous qualitative research which finds that countries with a high degree of financial development or digital infrastructure have more developed CBDC projects. Further, we obtain robust results when predicting the CBDCPI at different points in time.




\end{abstract}
%
%

\section{Introduction}

Recent advances in financial technology and distributed ledger techniques \cite{vonzurgathen2015,Narayanan2016} have paved the way to the extensive use of digital currencies. Although the advancement of these currencies came from private initiatives such as Bitcoin \cite{SatoshiNakamoto2008}, Ethereum \cite{wood2014ethereum}, and Libra~\cite{web:diem,Bruhl2019LIBRA}, researchers and  policymakers are contemplating whether central banks can also issue their own digital currencies, usually referred to as \emph{central bank digital currency} (CBDC). 

There has been a great deal of discussion about the implications of introducing CBDCs. Although the concept of a CBDC has existed for quite a long time~\cite{Tobin1987}, the stance toward whether central banks should introduce them has changed drastically over the past year. Initially the focus of central banks was on systemic implications \cite{Barontoni2019}  but several factors deriving from the benefits of the digital money have recently motivated central banks to issue a CBDC. This trend has further been fueled by the declining use of cash due to the growth of cashless payments, the possible introduction of global stablecoins \cite{FSB2020} and the Covid-19 pandemic \cite{BIScovid2020}.

In fact, a great number of central banks are undertaking extensive work on CBDC \cite{Boar2020}, several of which have issued research or statements on the related motivations, architectures, risks, and benefits. For instance, Boar et al. \cite{Boar2020} refer to the observed shift to intensive practical development from conceptual research, which is found in emerging markets, driven by stronger motivations than those of advanced economy central banks. In practice, several central banks issued their CBDCs between August and December 2020 (see Section~\ref{ssec:CBDCPI}). Further, a few central banks are aiming to issue their CBDCs in the next few years, which has attracted a lot of attention \cite{BIS_Auer2020}; for example, beginning of this year saw the closing of the ECB digital euro consultation with record level of public feedback \cite{ECB2021}. This move is consistent with the observation that the introduction of a CBDC can present a significant innovation in money and banking history \cite{FERNANDEZVILLAVERDE2020}.

However, despite the great amount of analysis conducted regarding important questions surrounding CBDCs, relatively little quantitative analysis has been undertaken especially on the drivers of the CBDC projects. The previous research include the potential risk and benefits of introducing the CBDC, and quantitatively, the welfare gains of introducing CBDC into the economy \cite{BoC2018}. Of the relevant research that exists in this vein, \cite{BIS_Auer2020} suggests by taking an ordered probit approach \cite{probit1975} for comprehensive cross country database that the majority of the CBDC projects are found in digitised economies with a high capacity for innovation. They conclude that some of the potential drivers of CBDC development are related to factors affecting a country’s digital infrastructure, innovation capacity, institutional quality, development and financial inclusion, public interest in CBDCs, and cross-border transactions.

This study examines the economic and institutional drivers of CBDC projects by applying machine learning techniques to the related variables obtained from official sources that are available for a wide cross section of countries.
Our primary objective is to improve understanding the dominant drivers for CBDCs and the factors that increases the possibly of a country to accelerate this effort.
We use the CBDC project index (CBDCPI) \cite{BIS_Auer2020} as our objective variable and factors affecting a country’s digital or technological capability and government effectiveness as independent variables, in order to reduce the problem to identifying the independent variables with the most predictive power.
To accomplish this, we utilise machine learning techniques to predict the CBDCPI and pick the most important variables for our model.

We compare two types of classifiers that are able to learn non-linear functions: a multilayer perceptron (MLP) \cite{Rosenblatt1961,MLP1986} and a random forest \cite{rf2001}.
In the experiment, we find random forest performs better than MLP, and that the financial development index \cite{Svirydzenka2016} is particularly the most important feature for our model, followed by the GDP per capita and the voice and accountability, when explaining the CBDCPI drivers for August 2020. This corresponds \cite{BIS_Auer2020}, which concluded that more developed CBDC projects can be found in countries with higher financial development index, digital infrastructure, GDP, and institutional characteristics. 
As a robustness check, we have performed the same analysis with full and aggregated data and with December 2020 CBDCPI. Results are broadly consistent, although there were some replacement in the ranks of important features.

This paper is structured as follows. Section 2 touches upon the preliminaries including the overview of Central bank digital currency (CBDC) and CBDC project index (CBDCPI). We subsequently presents the empirical models and data in Section 3, and then results in Section 4. We conclude in Section 5.


\section{Preliminaries}

In this section, we provide an overview of the key concepts from central bank digital currency (CBDC) relevant to this paper.

\subsection{Central bank digital currency (CBDC)}

Central bank digital currency (CBDC) is the digital form of fiat currency of a particular nation (or region). It differs from virtual currency and cryptocurrency, as CBDC is issued by the state and possesses the legal tender status declared by the government \cite{Grym2017}. Examples include the Digital Currency/Electronic Payments (DC/EP) by China’s central bank and e-krona by the central bank of Sweden.

The effects of introducing CBDC are receiving more attention than ever before. Although the concept of a CBDC was already proposed decades ago \cite{Tobin1987}, recent IT progress and its application to the financial industry have motivated central banks and academics to study the risks and merits of making CBDC accessible to the general public \cite{BIS2018,ecb20202351}, as presented in Table~\ref{tab:pro-con_CBDC} \cite{IMF_pro-con}. 

\begin{table}[htbp]\centering
\caption{Benefits and challenges of CBDCs}
\label{tab:pro-con_CBDC}
\begin{tabular}{llc}
\toprule
~\textbf{Advantages} & ~~\textbf{Disadvantages}\\
\hline
~Low cost of cash (efficient payment system) & ~~Higher run risk \\
~Enhances financial inclusion & ~~Disintermediation of commercial banks\\
~Stabilises the payment system & ~~Enhances currency substitution\\
~Faster transmission of monetary policy & ~~Risk and cost for central banks\\
\bottomrule
\end{tabular}
\end{table}

Further, attitudes towards whether CBDCs should be issued by central banks have changed drastically over the past year. This derived from the diminishing use of cash due to the rise of cashless payments, the possible entry of global stablecoins \cite{FSB2020} and the Covid-19 pandemic. Specifically, social distancing measures and public concerns that cash may transmit the Covid-19 virus and novel government-to-person payment schemes have further fueled the shift toward digital payments, and may act as a driver of CBDC projects \cite{BIScovid2020}. 

\begin{figure}[tbp]
\includegraphics[width=\textwidth]{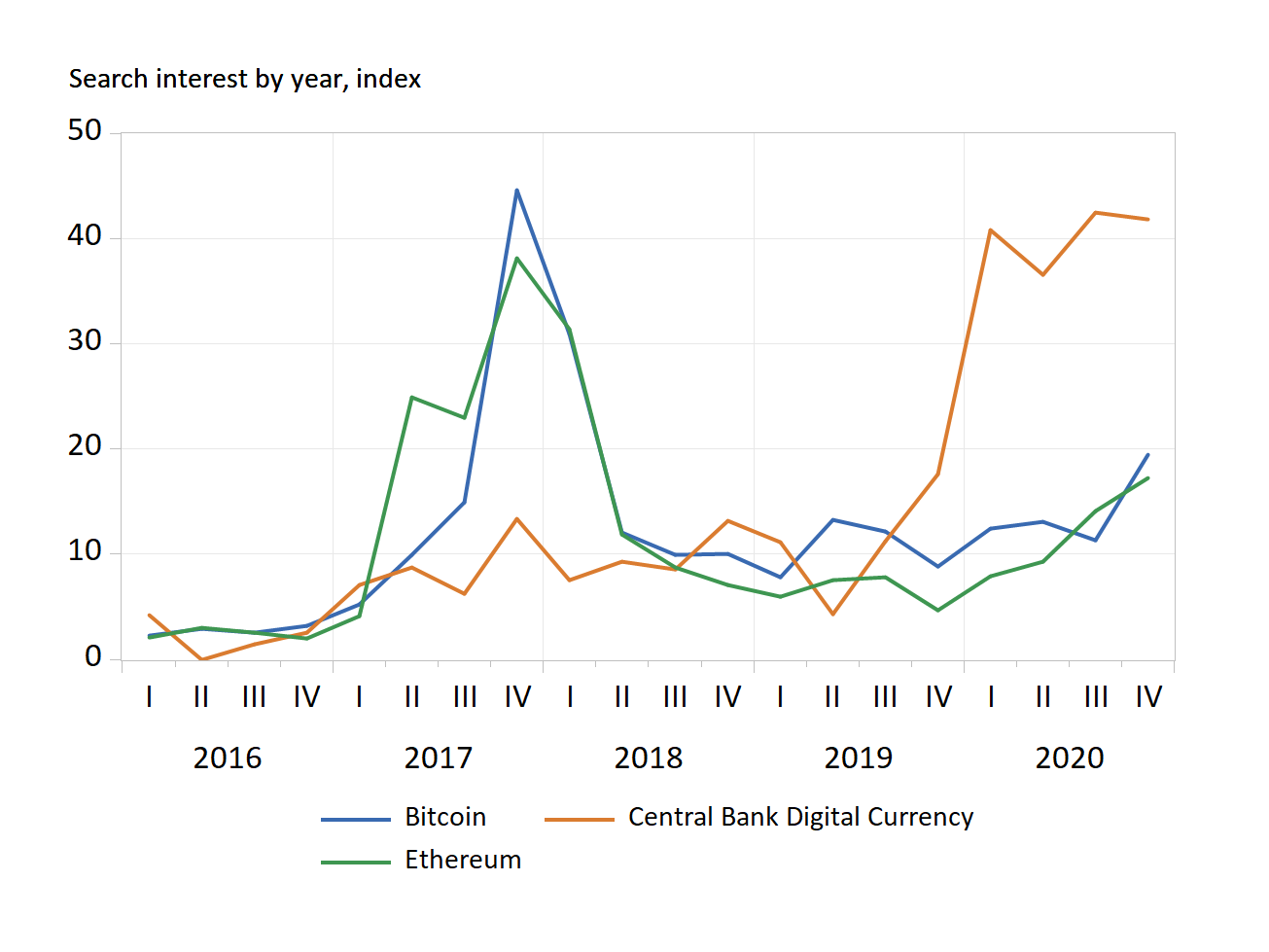}
\caption{Google Trends over time} \label{GoogleTrends}
\end{figure}

In fact, CBDCs have gained global attention, not only within central bank communities but also by the public. Fig.\ref{GoogleTrends} charts the Google search interest over time\footnote{We took 12-week moving average of Google Trends search results.}.
This shows that the current interest in CBDC is increasing, reaching a level almost as high as Bitcoin's during its price spike of 2017.

However, a majority of CBDCs are still in research or pilot stage, although a survey in early 2020 showed more than 80\% of central banks were studying the subject \cite{Economist2020}.
Further, the drivers of the CBDC projects are yet to be thoroughly investigated \cite{BIS_Auer2020}.

\subsection{CBDC project index (CBDCPI)}\label{ssec:CBDCPI}
The \textit{CBDC project index} (CBDCPI) was firstly proposed by Auer et al. \cite{BIS_Auer2020} to measure the central bank’s progress toward the development of a retail or wholesale CBDC. CBDCPI represents publicly announced work by central banks on CBDC related projects.
The index takes a value between 0 and 4 defined as follows:

0~~  - ~~No announced project\\
1~~  - ~~Public research studies\\
2~~  - ~~Ongoing or completed pilot\\
3~~  - ~~Live CBDC

There are two sub-indices, one for retail and one for wholesale CBDC projects.
Wholesale CBDC is devised as the new instrument for settlement between financial organisations, whereas retail CBDC aims to replace cash with the properties of central bank liability.
The overall index for a country is the maximum of these two sub-indices.

According to the dataset provided by the online annex of \cite{BIS_Auer2020}, there was no country with index 3 (live CBDC) as of August 2020 but by the end of 2020, seven countries and jurisdictions, including Bahamas, Canada, Switzerland, France, have shifted to 3 (See Appendix 1 for the CBDCPI scores as of December 2020.)\footnote{The updated CBDC projects status is available in an online annex of \cite{BIS_Auer2020} (See \url{https://www.bis.org/publ/work880.htm}).
The information is said to have collected through desk research and with the help of contacts at several individual central banks.}. This clearly reflects the recent trend and discussion that have brought the analysis about digital money and CBDC to the fore \cite{BIS_press2020}.

\section{Data and Methodology}

Our main goal is to understand better what are the main drivers for CBDCs and what makes a country more or less likely to push this effort.
Using the CBDC project index (CBDCPI) as our objective variable and factors affecting a country’s digital or technological capability and government effectiveness as explanatory variables, we can reduce this to finding the explanatory variables with the most predictive power.
To achieve this, we leverage machine learning techniques to predict the CBDCPI and extract the most important variables for our model.
We use the rest of this section to describe our data and provide details about our machine-learning based methodology.

\subsection{Data}\label{ssec:data}

We extract our dataset from the World Bank, the International Monetary Fund (IMF), and the data source of a BIS working paper \cite{BIS_Auer2020}.
It contains 16 variables, including data from 2000 to 2019, with more than 170 countries and jurisdictions.


Our variables can be divided into the following categories: i) digital infrastructure, ii) development and financial inclusion, iii) institutional characteristics, iv) innovation environment, v) demographic characteristics, and vi) cross-border transactions.
Each variable contains several years of data ranging from 2000 to 2019.
Although our dataset holds data about several financial development sub-indices, we only include the top-level index\footnote{See Financial Development Index Database  (https://data.imf.org/?sk=f8032e80-b36c-43b1-ac26-493c5b1cd33b) for more information.}.
We perform our analysis both using the \emph{full data} and an aggregated version of our data.
For the \emph{aggregated data}, we average each variable over the period 2014-2019, subject to data availability.
For further detail about the variables, see Appendix~\ref{sec:variables}.

The CBDCPI has 176 observations, one per country, each taking the value of 0, 1, 2, or 3, as described in Section~\ref{ssec:CBDCPI}.
We obtain the CDCPI for December 2020, in addition to its August 2020 value from previous research \cite{BIS_Auer2020}.
The major difference between these two variables is that, as of August 2020, there were no countries with a live CBDC (i.e. no index with value 3) while there were 7 of them with a live CBDC by December 2020.
We show the countries with a CDCPI of 3 as of December 2020 in Table~\ref{tab:top-cbdcpi}.

\begin{table}[tbp]
\caption{Countries with a CBDCPI of 3 as of December 2020.}\label{tab:top-cbdcpi}
\begin{tabular}{l|c c c c}
\toprule
\textbf{Country} &~~ \textbf{Overall*} &~~ \textbf{Overall (Aug 20)} &~~ \textbf{Retail*} &~~ \textbf{Wholesale*}\\ \hline
Bahamas &~~ 3 &~~ 2 &~~ 3 &~~ 0\\
Canada &~~	3 &~~ 2 &~~ 1 &~~ 2\\
Switzerland	&~~ 3 &~~ 1	&~~ 1 &~~ 2\\
Euro area (ECB)	&~~ 3	&~~ 2 &~~ 1 &~~ 2\\
France &~~ 3 &~~ 2	&~~ 1	&~~ 2\\
Japan &~~ 3 &~~ 2 &~~ 1 &~~ 2\\
South Africa &~~ 3 &~~ 1 &~~ 1 &~~ 2\\
\bottomrule
\addlinespace[1ex]
\multicolumn{3}{l}{*As of December 2020 (Source: online annex of \cite{BIS_Auer2020}).}
\end{tabular}
\end{table}

\subsection{Methodology}

Instead of the ordered probit approach \cite{probit1975} applied in \cite{BIS_Auer2020}, we model the problem as a classification task where the goal is to predict the CBDCPI given the set of input variables described above.
Given that the CBDCPI is a value between 0 and 3, it is easy to model as a categorical variable.

We settle on a random forest as our primary model, as it is known to be able to learn complex non-linear functions while being interpretable enough to extract the most important input variables \cite{rf2001}.
To obtain a point of comparison for the predictions of our random forest, we also train a multilayer perceptron \cite{Rosenblatt1961,MLP1986} on the same task.
However, given that multilayer perceptrons are not interpretable enough to understand the most important features, we only use these results for comparison.

We utilise this methodology with the full version and the aggregated version of the data to predict the CBDCPI both in August and December 2020.


\section{Results}
This section presents the results we obtained by training the models described above on our dataset.

Before starting our training process, we preprocess the data to filter out lacking data.
We first remove all the countries for which the CBDCPI is not available, as well as the countries for which one or more of the observed variables is not available (e.g. do not have a single year of data).
When a country is missing a year for a particular variable, we use the previous year to fill for it (e.g. if the GDP per capita is available for 2018 but not 2019, we set it to the value of 2019).

After this filtering process, we obtain a final list of 145 countries with only 6 countries having a CBDCPI of 3 as of December 2020; unfortunately, the Euro Area was not included in the final dataset as we did not have the financial development index data for it.
Further, we obtain a total of 13 variables for our aggregated data and 135 variables for the full data.

Then, we randomly split our dataset in two equal splits for training and testing.
We then tune the hyper-parameters of our two models.
We note that given the small size of our dataset, we use the full data instead of having a separate cross-validation set.
We find that our random forest works best with a total of 100 estimators.
For our multilayer perceptron, we use two layers, the first one with a number of neurons equal to the number of features and the second one a fixed size of 10 neurons.
Finally, we train the two models on our training data and evaluate them on our test data.
We present the accuracy of the two models in Table~\ref{tab:classification-results}.


\begin{table}[h]
    \caption{Accuracy of different classifiers on full and aggregated data}
    \label{tab:classification-results}    
    \centering
    \begin{tabular}{l l c c c c c}
    \toprule
            & ~~ &\multicolumn{2}{l}~~~{\bf Full data}~~~ &
            ~~& 
            \multicolumn{2}{l}{\bf Aggregated data}\\
            \cline{3-4}
            \cline{6-7}
    
    \bf ~~~&~~Classifier & ~~Train & Test && ~~Train & ~Test\\
    \midrule
    \multirow{2}{*}{Aug}~~ &
    \bf MLP & ~~1.0 & 0.79 && ~~0.99 & 0.74\\
    &\bf Random Forest & ~~1.0 & 0.78 && ~~1.0 & 0.77\\
    \cline{2-7}
    \addlinespace[1ex]
    \multirow{2}{*}{Dec}~~ &
    \bf MLP & ~~1.0 & 0.67 && ~~1.0 & 0.62\\
    &\bf Random Forest & ~~1.0 & 0.78 && ~~1.0 & 0.68\\
    \bottomrule
    \end{tabular}
\end{table}

Next, we use the features extracted by our random forest to understand better what the potential drivers of CBDC are.
We use the aggregated August 2020 data to compare with \cite{BIS_Auer2020} and see if the random forest achieves similar results as to the drivers for CBDCs.
We find that the financial development index~\cite{Svirydzenka2016} is by far the most important feature for our model, followed by the GDP per capita and the voice and accountability, when explaining the CBDCPI drivers for August 2020.
This is consistent with \cite{BIS_Auer2020}, stating that the CBDC projects to be more developed where there is higher financial development index, digital infrastructure, GDP, and institutional characteristics such as govenment effectiveness and voice and accountability.
We summarize the top 5 features and their importance in Table~\ref{tab:important-features}.

\begin{table}[tbp]
    \caption{Most important features for the random forest (Aug 2020, aggregated)}
    \label{tab:important-features}
    \setlength{\tabcolsep}{10pt}
    \centering
    \begin{tabular}{l c}
        \toprule
        \bf Feature & \bf Importance\\
        \midrule
        Financial Development Index & 0.165\\
        GDP per capita & 0.098\\
        Voice and accountability & 0.095\\
        Broadband subscriptions per 100 people & 0.093\\
        Government effectiveness & 0.087\\
        \bottomrule
    \end{tabular}
\end{table}

For robustness check, we conduct the same analysis with December 2020 CBDCPI data as an objective variable.
The results are consistent with the August CBDCPI data -- random forest performing much better and allows to extract the most important features used for classification.
The most significant features are: Mean 65+, FD index, Mobile cellular subscription, GDP per capita, and Voice and accountability, showing that the main features predicted as important are proved to be important with December 2020 CBDCPI data.
Further, similarly to the case of August 2020 data, we find that our model performs better with the full data rather than its aggregated version.

For the detailed results with the top 10 important features for both August and December 2020 CBDCPI with full and aggregated data, see Appendix~\ref{ssec:top-features} and \ref{ssec:top-features_full}.

\section{Conclusion}

In this paper, using a variety of machine learning methods used to learn complex non-linear functions, we investigated the importance of each economic and technological factors in predicting the progression of Central Bank Digital Currencies (CBDC) project within a country, using as our measure of this advancement the CBDC project index (CBDCPI).
We found that a financial development index is the most important feature for our model, followed by the GDP per capita and an index of the voice and accountability of the country’s population. 
Additionally, we confirmed that our results are in accordance with previous qualitative research which finds that countries with a high degree of financial development or digital infrastructure have more advanced CBDC projects. Moreover, we achieved robust results when examining the CBDCPI at different points in time.

\bibliographystyle{splncs04}
\bibliography{bibliography}
\newpage
\appendix
\section{Appendix}
This annex gives additional tables, regression results and figures to complement the paper. See main text for further discussion.

\subsection{CBDC projects status}

Below shows the part of the updated project score of global CBDC development efforts, relating to \cite{BIS_Auer2020} (as of December 2020)\footnote{The dataset includes all projects announced as of 1 December 2020. For more information, see \url{https://www.bis.org/publ/work880.htm}.}. Note that only the countries with index of 3 (live CBDC) and 2 (pilot) as of December 2020 are listed here.

\begin{table}[htbp]
\begin{tabular}{l|c c c c}
\toprule
\textbf{Country} &~~ \textbf{Overall*} &~~ \textbf{Overall (Aug 20)} &~~ \textbf{Retail*} &~~ \textbf{Wholesale*}\\ \hline
Bahamas &~~ 3 &~~ 2 &~~ 3 &~~ 0\\
Canada &~~	3 &~~ 2 &~~ 1 &~~ 2\\
Switzerland	&~~ 3 &~~ 1	&~~ 1 &~~ 2\\
Euro area (ECB)	&~~ 3	&~~ 2 &~~ 1 &~~ 2\\
France &~~ 3 &~~ 2	&~~ 1	&~~ 2\\
Japan &~~ 3 &~~ 2 &~~ 1 &~~ 2\\
South Africa &~~ 3 &~~ 1 &~~ 1 &~~ 2\\
United Arab Emirates~~ &~~ 2 &~~	2 &~~	0 &~~ 2\\
Australia &~~	2 &~~	1 &~~	1 &~~	1\\
China &~~	2 &~~	2 &~~	2 &~~	0\\
Ecuador &~~	2 &~~	2 &~~	2 &~~	0\\
Eastern Caribbean &~~	2 &~~	2 &~~	2 &~~ 0\\
United Kingdom &~~ 2 &~~ 2 &~~	1 &~~	1\\
Hong Kong &~~	2 &~~	2 &~~	0 &~~	2\\
Indonesia &~~	2 &~~	1 &~~	1 &~~	1\\
India &~~	2 &~~	0 &~~	1 &~~	1\\
South Korea &~~	2 &~~	2 &~~	2 &~~	0\\
Saudi Arabia &~~	2 &~~	2 &~~	0 &~~	2\\
Sweden &~~	2 &~~	2 &~~	2 &~~	0\\
Singapore &~~	2 &~~	2 &~~	0 &~~	2\\
Swaziland &~~	2 &~~	1 &~~	1 &~~	1\\
Thailand &~~	2 &~~	2 &~~	0 &~~	2\\
Ukraine &~~	2 &~~	2 &~~	2 &~~	0\\
Uruguay &~~	2 &~~	2&~~	2 &~~	0\\
\bottomrule
\addlinespace[1ex]
\multicolumn{3}{l}{*As of December 2020.}
\end{tabular}
\end{table}

\vspace{8mm}
\subsection{Table of observed variables}\label{sec:variables}
The following table lists the observed variables included in our analysis.
\begin{table}[h]
\begin{tabular}{l l c}
\toprule
\textbf{Variable} & ~~\textbf{Description} & \textbf{Source}\\ \hline
\addlinespace[1ex]
\textbf{Digital infrastructure}\\
Mobile subscriptions & ~~Mobile cellular subscriptions (per 100 people) & WB\\
Secure Internet & ~~Secure Internet servers (per 1 million people)  & WB\\
Fixed broadband & ~~Fixed broadband subscriptions (per 100 people) & WB\\
indiv. Internet use & ~~Individuals using the Internet (\% of population) & WB\\
\addlinespace[2ex]
\multicolumn{2}{l}{\textbf{Development and financial inclusion}}\\
Account ownership  &
\begin{tabular}{l}
~Account ownership at a financial institution or with a\\~mobile-money-service provider (\% of population ages 15+) 
\end{tabular}& WB\\
FD index & ~~Financial development index & IMF\cite{Svirydzenka2016}\\
GDP per capita & ~~GDP divided by midyear population (USD)  & WB\\
\addlinespace[2ex]
\multicolumn{2}{l}{\textbf{Institutional characteristics}}\\
Government effectiveness & ~~Quality of public services, policy implementation etc. & IMF\\
Regulatory quality & ~~Ability to formulate and implement sound policies etc. & IMF\\
Voice and accountability & ~~Extent of citizens' participation and freedom expression & IMF\\

\addlinespace[2ex]
\multicolumn{2}{l}{\textbf{Innovation environment}}\\
Access to electricity & ~~Access to electricity (\% of population) & WB\\

\addlinespace[2ex]
\multicolumn{2}{l}{\textbf{Demographic characteristics}}\\
\% of people over 65
 & ~~Total population 65 years of age or older & WB \\
 
\addlinespace[2ex]
\multicolumn{2}{l}{\textbf{Cross-border transactions}}\\
Trade (\% of GDP) & ~~Sum of exports and imports (\% of GDP) & WB \\


\bottomrule
\addlinespace[1ex]
\multicolumn{3}{l}{Source: World Bank, IMF, BIS (online annex of \cite{BIS_Auer2020}).} 
\end{tabular}
\end{table}


\newpage
\subsection{Top 10 features for the random forest classifier with aggregated data}
\label{ssec:top-features}
Table~\ref{tab:important-features_2008} and \ref{tab:important-features_2012} give the 10 most important independent variables for the random forest classifier with aggregated data (data averaged over the period 2014–19, subject to data availability), with August 2020 and December 2020 CBDCPI data as an objective variable, respectively.
\begin{table}[h]
    \caption{Most important features for the random forest classifier (Aug 2020)}
    \label{tab:important-features_2008}
    \setlength{\tabcolsep}{10pt}
    \centering
    \begin{tabular}{l c}
        \toprule
        \bf Feature & \bf Importance\\
        \midrule
        Financial Development Index & 0.165\\
        GDP per capita & 0.098\\
        Voice and accountability & 0.095\\
        Broadband subscriptions (per 100 people) & 0.093\\
        Government effectiveness & 0.087\\
        \% of people over 65 & 0.082\\
        Individuals using the Internet (\% of population) & 0.080\\
        Trade (\% of GDP) & 0.078\\
        Secure Internet servers (per 1 million people) & 0.072\\
        Regulatory quality & 0.068\\
        \bottomrule
    \end{tabular}
\end{table}

\begin{table}[h]
    \caption{Most important features for the random forest classifier (Dec 2020)}
    \label{tab:important-features_2012}
    \setlength{\tabcolsep}{10pt}
    \centering
    \begin{tabular}{l c}
        \toprule
        \bf Feature & \bf Importance\\
        \midrule
        \% of people over 65 & 0.134\\
        Financial Development Index & 0.109\\
        Mobile cellular subscriptions (per 100 people) & 0.101\\
        GDP per capita & 0.090\\
        Voice and accountability & 0.089\\  Secure Internet servers (per 1 million people) & 0.086\\   
        Individuals using the Internet (\% of population) & 0.080\\
        Government effectiveness & 0.080\\
        Broadband subscriptions (per 100 people) & 0.076\\
        Trade (\% of GDP) & 0.067\\        
        \bottomrule
    \end{tabular}
\end{table}

\newpage
\subsection{Top 10 features for the random forest classifier with full data}
\label{ssec:top-features_full}
Table~\ref{tab:important-features_full2008} and \ref{tab:important-features_full2012} show the 10 most important index for the random forest classifier with full data, with August 2020 and December 2020 CBDCPI data as an objective variable, respectively.
\begin{table}[h]
    \caption{Most important features for the random forest classifier (Aug 2020)}
    \label{tab:important-features_full2008}
    \setlength{\tabcolsep}{10pt}
    \centering
    \begin{tabular}{l c}
        \toprule
        \bf Feature & \bf Importance\\
        \midrule
        Financial Development Index & 0.052\\
        Government effectiveness [YR2019] & 0.033\\
        Government effectiveness [YR2018] & 0.020\\
        Broadband subscriptions (per 100 people) [YR2017] & 0.020\\
        Individuals using the Internet (\% of population) [YR2012]& 0.020\\
        Individuals using the Internet (\% of population) [YR2015]& 0.020\\
        GDP per capita [YR2016] & 0.018\\
        Government effectiveness [YR2015] & 0.016\\
        Mobile cellular subscriptions (per 100 people) [YR2016]& 0.015\\
        \% of people over 65 [YR1990]& 0.014\\
        \bottomrule
    \end{tabular}
\end{table}

\begin{table}[h]
    \caption{Most important features for the random forest classifier (Dec 2020)}
    \label{tab:important-features_full2012}
    \setlength{\tabcolsep}{10pt}
    \centering
    \begin{tabular}{l c}
        \toprule
        \bf Feature & \bf Importance\\
        \midrule
        Financial Development Index & 0.037\\
        \% of people over 65 [YR1990] & 0.035\\
        Mobile cellular subscriptions (per 100 people) [YR2019] & 0.025\\
        \% of people over 65 [YR2018] & 0.023\\
        \% of people over 65 [YR2017] & 0.021\\
        \% of people over 65 [YR2000] & 0.020\\
        Government effectiveness [YR2017] & 0.020\\
        \% of people over 65 [YR2015] & 0.020\\
        \% of people over 65 [YR2013] & 0.019\\
        Mobile cellular subscriptions (per 100 people) [YR2015] & 0.018\\       
        \bottomrule
    \end{tabular}
\end{table}

\end{document}